# Gender Prediction Based on Vietnamese Names with Machine Learning Techniques

Huy Quoc To
University of Information Technology,
Ho Chi Minh City, Vietnam
Vietnam National University, Ho Chi Minh City, Vietnam
huytq@uit.edu.vn

Kiet Van Nguyen
University of Information Technology,
Ho Chi Minh City, Vietnam
Vietnam National University, Ho Chi Minh City, Vietnam
kietnv@uit.edu.vn

Ngan Luu-Thuy Nguyen
University of Information Technology,
Ho Chi Minh City, Vietnam
Vietnam National University, Ho Chi Minh City, Vietnam
ngannlt@uit.edu.vn

Anh Gia-Tuan Nguyen
University of Information Technology,
Ho Chi Minh City, Vietnam
Vietnam National University, Ho Chi Minh City, Vietnam
anhngt@uit.edu.vn

## ABSTRACT
As biological gender is one of the aspects of presenting individual human, much work has been done on gender classification based on people names. The proposals for English and Chinese languages are tremendous; still, there have been few works done for Vietnamese so far. We propose a new dataset for gender prediction based on Vietnamese names. This dataset comprises over 26,000 full names annotated with genders. This dataset is available on our website for research purposes. In addition, this paper describes six machine learning algorithms (Support Vector Machine, Multinomial Naive Bayes, Bernoulli Naive Bayes, Decision Tree, Random Forrest and Logistic Regression) and a deep learning model (LSTM) with fastText word embedding for gender prediction on Vietnamese names. We create a dataset and investigate the impact of each name component on detecting gender. As a result, the best F1-score that we have achieved is up to 96% on LSTM model and we generate a web API based on our trained model.

## 1. INTRODUCTION
Gender prediction based on human's names is a topic on which a vast number of researches are investigating [1-5]. There have been large numbers of investments on identifying the most suitable algorithms and models for different languages. The reason behind for these researches is the potential benefits of predicting two common genders (Male and Female). Consequently, there are a variety of online web applications and tools that are capable of pointing out the gender from user's first name [6 -9].

These existing gender predicting systems are useful and practical in terms of third-party APIs providers for other applications [5]. A typical utilization is online registration forms and documents. For instance, giving a form where the users have to fill up their names and genders, the application will automatically choose the corresponding gender field after the users have typed in their names. Predicting gender eliminates the amount of time which the users need to away from keyboard and click on computer mouse or track-pad to check on the field. It enhances the overall user experience while using the services.

In addition, the co-reference resolution is a Natural Language Processing task to which gender classification is possible to be applied. The co-reference mission is to identify the same entity that all the objects in one text are referring. For example, "James is having dinner with Mia. He is having beef steak.". In this scenario, He is referring to James which is a Male. Therefore, determining gender of a given name is sufficiently essential to mark references to correct entity [10]. On the other hand, not many related researches on Vietnamese names have been done over the years. Thus, the goal of this paper is to perform gender classification on Vietnamese names using machine learning models. From there, we can study further on the naming conventions and how efficient the machine learning algorithm predicts the genders using Vietnamese names.

In this paper, we have three main contributions described as follows:

- First, we build a new dataset for analysing Vietnamese names called UIT-ViNames. The dataset consists of 26,850 Vietnamese full names annotated with 1 for male and 0 for female. UIT-ViNames is available for research purposes at our website. [1]
- Second, we experiment new implementations and comparison of multiple traditional machine learning and deep learning models on predicting genders on our dataset. As a result, LSTM model with fastText word embedding achieves the highest average score with 95.89%.
- Lastly, we perform analyses on Vietnamese names which provides a deeper understanding on naming convention and increases the efficiency of using traditional machine learning and deep learning for gender prediction.

In Section II, we perform literature review on previous studies. Then in Section III, we describe the our dataset and how we collect it. Next, we present a detailed description on our approach for this topic in Section IV and our experiments in Section V. Finally, in Section VI we draw an overall conclusion and several future works.

## 2. RELATED WORK

[1] https://sites.google.com/uit.edu.vn/uit-nlp/

In previous related works, there are several approaches for English languages, while there are few proposals for Vietnamese. Koppel et al. [2] provided a methodology on identifying genders of written document's authors. They combined syntactic and lexical features on the subset of British National Scopus (BNC) dataset. The experimental result was about 80% which was the outcome of Part of Speech (POS) n-grams.

Peersman et al. [4] described an approach for classifying genders on social networks. The authors collected the data from Netlog which is a Dutch social network. They performed SVM technique for this short text classification task. The highest accuracy score was 88.8% when applying 50,000 most information word unigrams.

In terms of tools, GenderAPI [6] and NameAPI [7] are the most common used among the community. However, their drawback is obvious which includes limited dictionary and not open-sourced. In addition, the prediction accuracy on languages such as Chinese and Russian is fairly low. Zhao and Kamareddine [5] proposed an advance application based on existing tools within the UK, Malaysia and China. The average performance of their prediction system was 96.5%.

In more recent study, Jia and Zhao [1] described a novel approach utilizing fastText word embedding on BERT-based model to classify genders based on Chinese names. In their paper, they provided a comparison on traditional models, namely Naive Bayes, Gradient Boosting Decision Tree (GBDT) and Random Forest. As the result, they achieved 93.45% test accuracy on BERT-based model.

Panchenko and Teterin [11] proposed an approach on Russian names for gender classification task using linear supervised model. The experiment was performed on a dataset of 100,000 full names from Facebook using n-grams, word embedding and dictionary features. The authors reported that the model performance reached up to 96% of accuracy.

As the languages differentiate among nations and regions, a study on the component of names are necessary to further understand how their impacts on genders. Since Chinese is a logo-syllabic language, the approaches for this task are focusing on the character itself. In contrast, Vietnamese language is based on Latin alphabet which is similar to English. Therefore we only consider and compare the characteristic Vietnamese names to English names. Le [12] defined that Vietnamese name has three components, respectively Họ (surname/family name), Tên Đệm (middle name) and Tên (given/first name). According to Le's research, the official order of Vietnamese name is differ from English names. Specifically, in English, the order of name component is first name, middle name, and family name (surname). For example, John Doe consists of John (given/first name) and Doe (family name). In a comparison, family name is placed as last component in English name, while in Vietnamese name, it is place at first position. In next section, we describe the data collection procedure and analyze the dataset in details.

# 3. DATASET

In this section, we present how we collect the data and analyze it distributions and characteristics. Our method of dividing Vietnamese name into components is derived from Le's study [12] since it provides a baseline for investigating Vietnamese names. We evaluate each individual component in order to determine its weight of efficiency in identifying genders.

## 3.1 Dataset Collection

Our data is collected from several universities in Vietnam. The students in those universities come from provinces and cities across Vietnam, so that it diversifies the collection of data and covers a wider range of names. Additionally, the data only contains names and genders which avoids revealing the identities of the students. The name field includes full names of the students which are annotated with two labels (Male and Female). A capture of names of males and females is displayed in Table 1. Our dataset consists of 26,850 samples of which proportion of two labels is fairly balance. The percentage of male and female, respectively, is 57.71% and 42.29%.

**Table 1. Examples for male and female names.**

| No. | Full name | Gender |
|---|---|---|
| **1** | Võ Minh Đủ (Vo Minh Du) | 1 |
| **2** | Nguyễn Thị Hiền (Nguyen Thi Hien) | 0 |

## 3.2 Dataset Analysis

In the context of recognizing individual genders, there are evidences showing that people's last names have a lower impact than first names (including middle name) [13]. We also compare the distributions of last names in male's comparing to female's for our dataset which are shown in Figure 1.

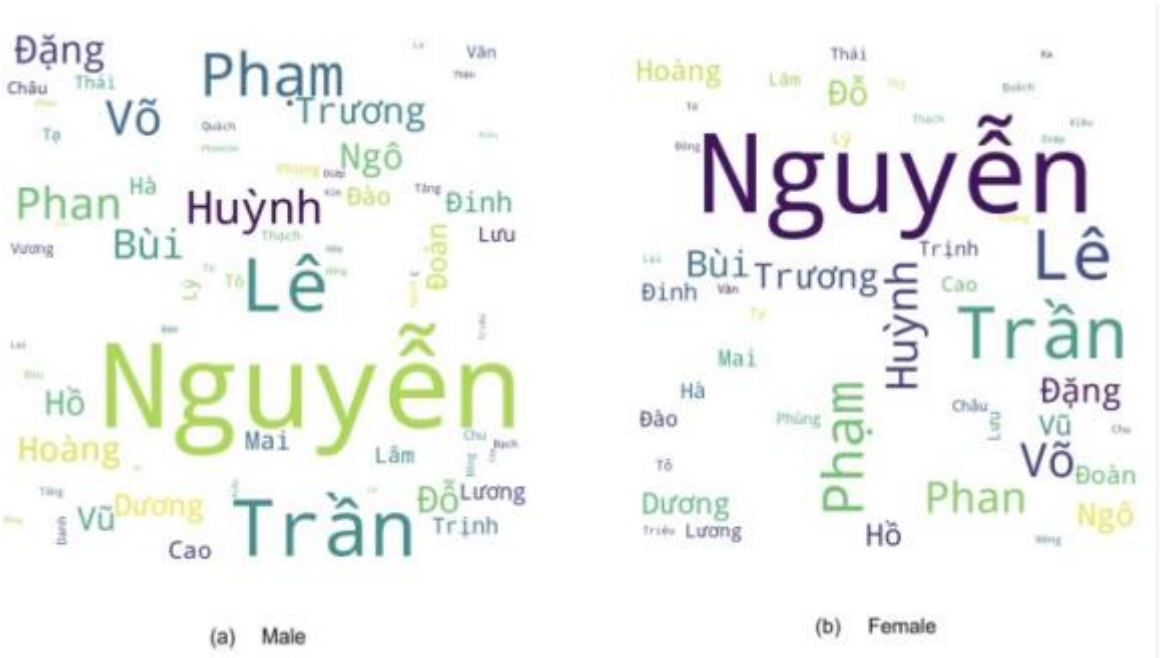

**Figure 1. The distribution male and female last names in Vietnamese.**

Figure 1 illustrates that the most common Vietnamese last names for both males and females are identical. This is a concrete evidence for the fact that, in Vietnamese names, last name has zero or very low impact on determining genders. Hence, we only consider first names and middle names in the gender predictions. The distribution of first names of Vietnamese males and females is compared in Figure 3. In comparison, those figures demonstrating the common Vietnamese first names have the words density that are significantly higher than of which the Vietnamese last names. It is clearly shown that **Nguyễn** (Nguyen) is the most usual Vietnamese last names. There are researches showing that approximately 40% of Vietnamese's last name is **Nguyen** which reflects in our dataset as well. Based on Figure 1, it also indicates the fact that the variety of last names in Vietnamese is fairly small. This is clearly demonstrated by the densities of both figures for last names. Figure 2 shows that majority of Vietnamese male's middle names are **Minh** (Minh) and Văn (Van); which on the other side, Thị (Thi) is widely applied for naming females.

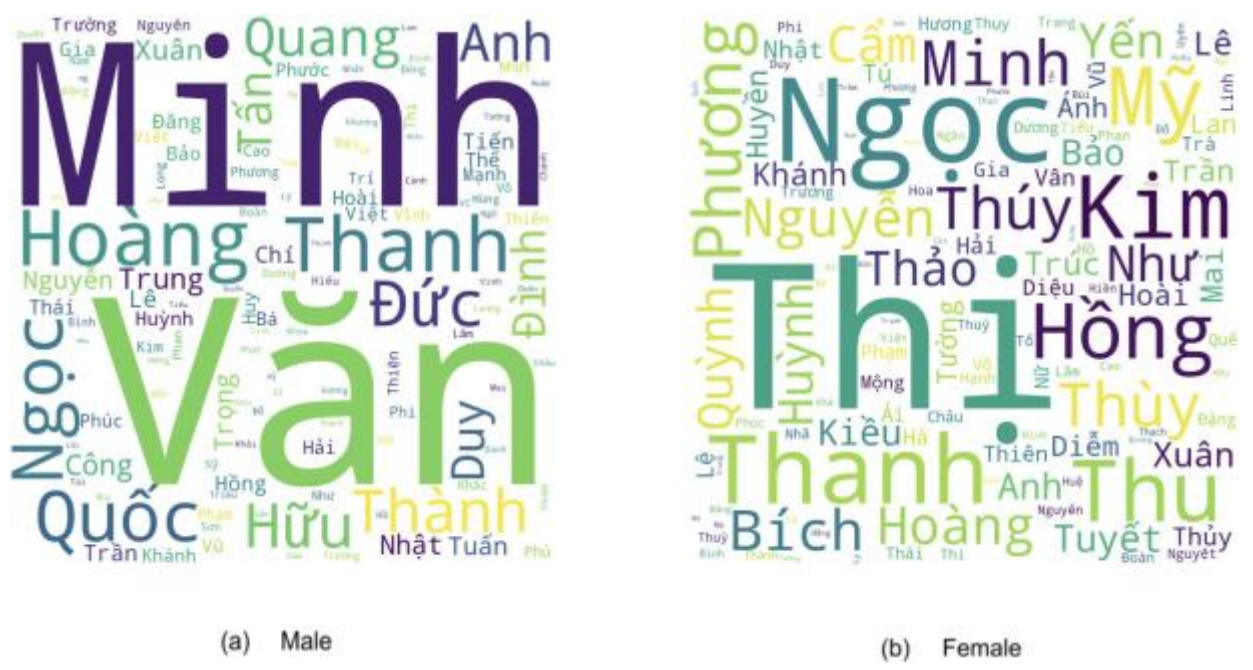

**Figure 2. The distribution of male and female middle names in Vietnamese.**

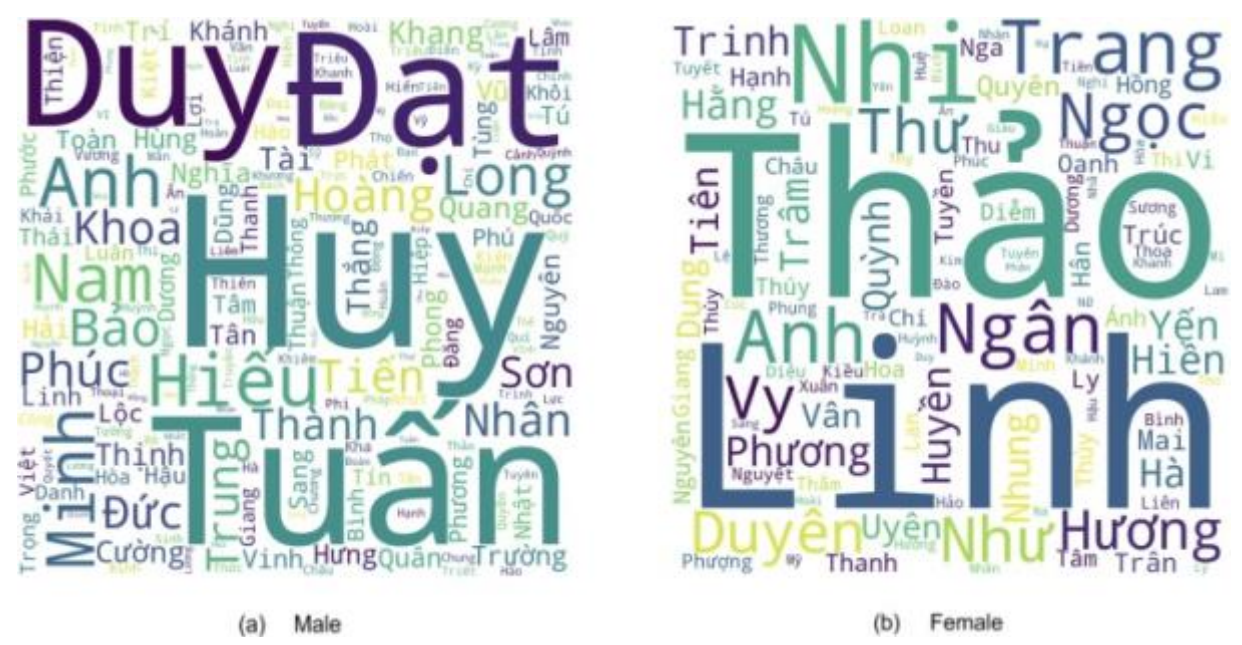

**Figure 3. The distribution of male and female first names in Vietnamese. The density is based on the diversity of words in the dataset. The size of word is calculated by its number of appearance.**

Furthermore, Figure 3 illustrates some of the most common first names of Vietnamese males and females. In particularly, it points out that **Anh** is the one that appears frequently in both Male and Female names. For example, **Tuấn Anh** (Tuan Anh) is more likely a **Male** while ***Tú Anh*** (Tu Anh) has higher percentage to be a ***Female***. This observation suggests that a simple first name is not enough to robustly classify the genders. Therefore, middle names are necessary in order to rank up the possibility of detecting genders correctly.

# 4. OUR APPROACHES

In this section, we present a description on the research methods implemented in this paper as well as the reason why we come up with the decisions on choosing models. We define this problem as a binary classification task since there are two labels (Male and Female) needed for this task. The purpose of this paper is to figure out the most efficient machine learning technique for detecting genders based on names, therefore, we implement commonly used models which namely are Naive Bayes Bernoulli, Naive Bayes Multinomial, Support Vector Machine, Logistic Regression, Decision Tree, Random Forest and deep learning model Long short-term memory.

## 4.1 Pre-processing Data

The data collected has been gone through two steps of preprocessing. Firstly, we cut off all the last names and those are stored separately in two dictionaries which are male's last name and female's last name. We have discovered that last names have inconsiderable influence on genders which is also clearly stated in Section III. Next, we perform cleaning data by removing special characters, double spacing, and correct the misspellings. We also convert the names into lower case in order to strengthen the integrity for our data. Data deduplication is an important step that excludes all the duplicates in our dataset. Consequently, we obtain a set of full names with no extra spacing and special characters.

## 4.2 Gender-prediction Models

### *4.1.1 Traditional machine learning models*

Naive Bayes is a widely applied for text classifier [14], and especially, gender predictions using English and Chinese names [5].

Multinomial (Multinomial NB) is specially design for text classification by identifying the appearance frequency of a specific words in Vietnamese names having higher likelihood of presence in one gender. For instance, Thị (Thi) is used for naming females while Văn (Van) is commonly in a male's name. Thus, Multinomia model is a decent choice for this classification task. Multivariate Bernoulli Naive Bayes Classifier (BNB) model is moderately similar to the Multinomial algorithm in terms of classification process. In comparison, BNB model focuses on the present state of a word on the documents under consideration while Multinomial model approach, as mentioned, considers the term frequency in documents [15].

We utilize Support Vector Machine (SVM) because it provides a superior method for text classification using kernel function to handle nonlinear spaces [16]. The Vietnamese names have some words that both frequently appear in two labels, thus, SVM algorithm in this scenario is able to provide a better results on predictions.

Logistic Regression (LR) is another discriminative algorithm that we would like to investigate its capability of determining genders using Vietnamese names. Logistic regression could be applied for text classification task since its usage is to express the relationship between the dependent binary variables and independent variables [17].

Decision Tree (DT) is a classification technique that represents each feasible outcome into each possible result using branching method [18]. Farooqui et al [19] clearly stated that DT's performance is much higher when the classification involved decision making. In our task, each individual word in a name has a weighted influence towards one gender.

Random Forest (RF) is an ensemble algorithm that comprises multiple decision tress. We use this uncorrelated model to compare with DT model. Maruf et al. [20] described a new method on Random Forest and Feature Selection (FS) for text classification and achieved macro-F1 score 73% higher than normal FS algorithm.

### *4.1.2 Deep learning model*

Long short-term memory (LSTM) is a deep learning model whose architecture includes gates and memory cells. These components allow LSTM to store and retrieve the information through operations such as write, read and reset. Johnson and Zhang [21] investigated one-hot LSTM with region embedding which yields the effectiveness of the approach in categorizing text. We implement this method with fastText word embedding to measure the performance of a deep learning algorithm comparing to traditional machine learning techniques.

## 5. EXPERIMENTS

In this Section, we describe the process of setting up the experiments using NB Bernoulli, Naive Bayes Multinomial, Support Vector Machine, Logistic Regression, and Long Shortterm Memory as well as examining the outputs of each technique by investigating the confusion matrices and wrong predictions. There are three main steps which are data preparation, experiment configurations and result analysis.

### 5.1 Data Preparation

First of all, we randomly divide our dataset into three different subsets: training, development, and test. In our dataset, the proportion of male and female is relatively balance. Thus, each of the subset is designed to maintain this equity and avoid bias against any side. We separate the corpus in the proportion of 70%, 10%, and 20%, for training set, development set, and test set respectively. We start by individually separating the data into training, development and test subset for each gender label. Then, we merge two subsets by their gender label, for example, train set for male is combined with train set for female. After the data has been splitted up, the names (features) are needed to be converted into individual vector. In our experiments, we evaluate **Count Vector** and **TF-IDF** features for vectorizing and tokenizing the data.

### 5.2 Experimental Settings

Since not many related researches for this task have been published, we based on other previous experiments on text classification to set up our parameters for each model. When tokenizing the names, we also ignore configuring the threshold for maximum and minimum frequently of words. The reason is that the stop words are out of concern so it is not necessary to limit out the repetition. For Count Vector, we use the baseline configuration but for TF-IDF, we set *maximum features = 4,000*.

After each run, we capture the precision and recall to calculate F1-score as the final results; as well as the confusion matrix to further investigate wrong predictions. Since the results of binary classification task can be divided into four classes: True Positive (TP), False Positive (FP), True Negative (TN) and False Negative (FN), we apply macro average scoring method to calculate F1-score.

On the other hand, in this paper, we apply fastText word embedding for LSTM model as it supports various languages. Grave et al. [22] used fastText for word representation in 157 languages. We set embedding size as *300* and configure *batch size = 32* and *epoch = 2* in our experiment.

### 5.3 Result Analysis

We report the final results of four machine learning algorithms for both tokenizing methods in Table 2. The results visually specify that Support Vector Machine produces the best results with count vector method while using TF-IDF. Whereas, Bernoulli Naive Bayes reaches the highest performance among other traditional machine learning methods. However, Table 3 yields that deep learning model LSTM reaches the best F1-score of 95.89%. Specifically, we record the best precision, recall and F1-score of all traditional machine learning models in comparison with LSTM. As a result, the deep learning model LSTM achieve the best experimental performance on our dataset. We notice that the results in both cases are approximately equivalent. By inspecting the wrong predictions, we discover that the similarity in wrong classification results. The term frequency of name component is the factor of this confusion. For instance, Figure 2 displays that the words **Ngọc** (Ngoc) or **Thanh** (Thanh) frequently appear in Vietnamese female’s and male’s middle names.

We also conduct ablation experiments in which we systematically remove name components of the input to understand their impact on Vietnamese gender prediction. In this examination, we only utilize SVM with Count Vector, Bernoulli Naive Bayes with TF-IDF and LSTM since they give the best outputs in this task. Table 4 is the results when we run the test on seven (7) name combinations. They indicate the fact that the concatenation of middle name and first name outperforms the others. The results again strengthen the proclamation that Vietnamese surnames have remarkably low impact on detecting genders. This experiment also yields the importance of Vietnamese middle names in gender detection task. Its achievements among other standalone name components is highest with the average of 91.35% using SVM with Count Vector, 90.94% using Bernoulli Naive Bayes with TF-IDF and 92.07% on LSTM.

**Table 2. F1-score of 6 machine learning techniques.**

| Models | Count Vector | | | TF-IDF | | |
|---|---|---|---|---|---|---|
| | Male | Female | Average | Male | Female | Average |
| **Multinomial NB** | 95.92 | 94.41 | 95.16 | 95.16 | 93.3 | 94.23 |
| **Bernoulli NB** | 96.03 | 94.46 | 95.25 | **96.06** | **94.49** | **95.28** |
| **Logistic Regression** | 95.96 | 94.32 | 95.14 | 95.75 | 94.03 | 94.89 |
| **Support Vector Machine** | **96.06** | **94.49** | **95.28** | 95.75 | 94.20 | 94.89 |
| **Decision Tree** | 95.07 | 93.04 | 94.05 | 94.32 | 92.26 | 93.78 |
| **Random Forest** | 95.17 | 93.38 | 94.28 | 95.65 | 94.03 | 94.84 |

**Table 3. Experimental performance of examined models on our dataset.**

| Model | Precision | Recall | F1-score |
|---|---|---|---|

| | | | | |
|---|---|---|---|---|
| | **Multinomial NB** | 95.28 | 95.06 | 95.16 |
| | **Logistic Regression** | 95.50 | 94.87 | 95.14 |
| **Traditional Machine Learning** | **Bernoulli NB** | 95.59 | 95.03 | 95.28 |
| | **Support Vector Machine** | 95.57 | 95.05 | 95.28 |
| | **Decision Tree** | 94.44 | 93.77 | 94.05 |
| | **Random Forest** | 94.98 | 94.73 | 94.87 |
| **Deep Learning** | **LSTM** | **96.11** | **95.70** | **95.89** |

**Table 4. F1-score of different combinations of name components.**

| **Name Components** | **SVM + Count Vector** | | | **Bernoulli NB + TF-IDF** | | | **LSTM** | | |
|---|---|---|---|---|---|---|---|---|---|
| | **Male** | **Female** | **Average** | **Male** | **Female** | **Average** | **Male** | **Female** | **Average** |
| **Family Name (FaN)** | 79.89 | 5.31 | 39.60 | 73.95 | 4.49 | 39.22 | 73.29 | 3.16 | 38.23 |
| **Middle Name (MN)** | 93.04 | 89.66 | 91.35 | 92.70 | 89.18 | 90.94 | 91.87. | 87.35 | 89.61 |
| **First Name (FiN)** | 89.37 | 84.71 | 87.04 | 89.35 | 84.48 | 86.92 | 86.26 | 73.79 | 80.02 |
| **FaN + MN** | 92.86 | 89.20 | 91.03 | 92.44 | 88.19 | 90.32 | 91.72 | 87.89. | 89.80 |
| **FaN + FiN** | 89.52 | 85.50 | 87.51 | 89.36 | 85.00 | 87.19 | 88.43 | 81.36 | 84.90 |
| **MN + FiN** | **96.06** | **94.49** | **95.28** | **96.06** | **94.49** | **95.28** | **96.56** | **95.22** | **95.89** |
| **FaN + MN + FiN** | 95.47 | 93.77 | 94.62 | 95.68 | 94.08 | 94.88 | 95.67 | 94.43. | 95.05 |

**Table 5. Common Wrong Predictions.**

| **Middle and First Name** | **True Gender** | **Prediction** |
|---|---|---|
| Đức Phương (Duc Phuong) | | |
| Đăng Tú (Dang Tu) | 1 | 0 |
| Lâm Anh (Lam Anh) | | |
| Thục Vy (Thuc Vy) | | |
| Uyên Quang ( Uyen Quang) | | |
| Gia Hảo (Gia Hao) | | |
| Minh Tuyên (Minh Tuyen) | 0 | 1 |
| Tú Văn (Tu Van) | | |
| Minh Ngọc (Minh Ngoc) | | |
| Ý Duy (Y Duy) | | |

Table 5 is a list of examples of misclassification for two labels which we to further analyze the factors causing errors. In both cases, we notice that there are some words appear in male names which are normally chosen when naming females and vice versa. For example, Phương (Phuong), Tú (Tu), Lâm Anh (Lam Anh), Thục Vy (Thuc Vy), and Uyên (Uyen) are regularly in female names, however in our dataset, they are in male names. In contrast, Hảo (Hao), Tuyên (Tuyen), Văn (Van), Minh (Minh), and Duy (Duy) are typically male's names but not female's.

# 6. CONCLUSION AND FUTURE WORK

In this paper, we have presented several approaches for gender classification task based on Vietnamese names. Our experiment is conducted on six (6) traditional machine learning techniques and one (1) deep learning model. LSTM deep learning model reaches excellent results on our UIT-ViNames dataset which is up to 96%. Hence, we also generate a simple web API service based on this LSTM pre-trained model. In our experiment, we also show the importance of Vietnamese middle name when detecting gender.

In conclusion, the analyses on our dataset and experimental results show that Vietnamese family names have low influence on genders, while middle names are the most important as well as LSTM model produces the best F1-score on gender detection task. In future work, we try to solve the issues of wrong prediction on majority of female names. We also plan to expand the name diversity in our dataset and study the efficiency of transfer learning models such as BERT and other deep learning models on this task.

# 7. ACKNOWLEDGEMENT

This research is funded by University of Information Technology-Vietnam National University HoChiMinh City under grant number D1-2020-22.